\documentclass{article}

\pdfoutput=1
\usepackage{PRIMEarxiv}

\usepackage[utf8]{inputenc} 
\usepackage[T1]{fontenc}    
\usepackage{hyperref}       
\usepackage{url}            
\usepackage{booktabs}       
\usepackage{amsfonts}       
\usepackage{nicefrac}       
\usepackage{microtype}      
\usepackage{lipsum}
\usepackage{fancyhdr}       
\usepackage{graphicx}       
\graphicspath{{media/}}     
\usepackage{amssymb}
\usepackage{amsmath}
\usepackage{bm}
\usepackage{xcolor}
\usepackage{setspace}
\usepackage{authblk} 
\usepackage{subcaption}
\usepackage{lineno,bbm,xcolor,multirow,pifont}

\author[1]{Xi Chen}
\author[2]{Rahul Bhadani}
\author[3]{Zhanbo Sun}
\author[1]{Larry Head}
\affil[1]{The University of Arizona, Tucson, USA}
\affil[2]{The University of Alabama in Huntsville, Huntsville, USA}
\affil[3]{Southwest Jiaotong University, Chengdu, China}
  
\title{MSMA: Multi-agent Trajectory Prediction in Connected and Autonomous Vehicle Environment with Multi-source Data Integration
\thanks{\textit{\underline{Citation}}: 
\textbf{Authors. Title. Pages.... DOI:000000/11111.}} 
}

\begin{document}
\maketitle

\begin{abstract}
The prediction of surrounding vehicle trajectories is crucial for collision-free path planning. In this study, we focus on a scenario where a connected and autonomous vehicle (CAV) serves as the central agent, utilizing both sensors and communication technologies to perceive its surrounding traffics consisting of autonomous vehicles (AVs), connected vehicles (CVs), and human-driven vehicles (HDVs). Our trajectory prediction task is aimed at all the detected surrounding vehicles. To effectively integrate the multi-source data from both sensor and communication technologies, we propose a deep learning framework called MSMA utilizing a cross-attention module for multi-source data fusion. Vector map data is utilized to provide contextual information. The trajectory dataset is collected in CARLA simulator with synthesized data errors introduced. Numerical experiments demonstrate that in a mixed traffic flow scenario, the integration of data from different sources enhances our understanding of the environment. This notably improves trajectory prediction accuracy, particularly in situations with a high CV market penetration rate. The code is available at: \url{https://github.com/xichennn/MSMA}.
\end{abstract}


\section{Introduction}
\label{sec:intro}

\begin{figure}
  \begin{subfigure}{0.45\textwidth}
    \centering
    \includegraphics[width=\textwidth]{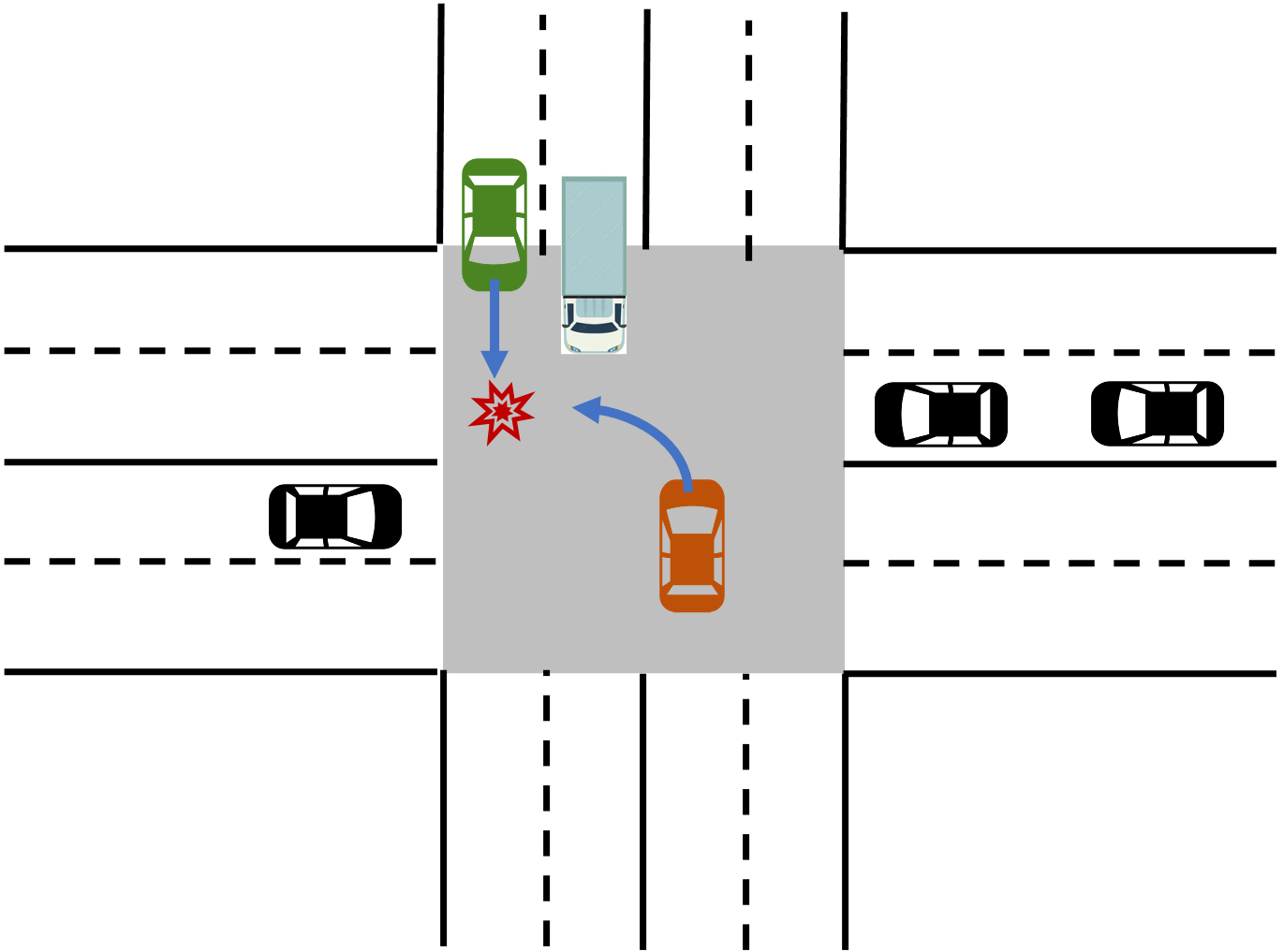} 
    \caption{ Left-turn collision}
  \end{subfigure}
  \hfill
  \begin{subfigure}{0.45\textwidth}
    \centering
    \includegraphics[width=\textwidth]{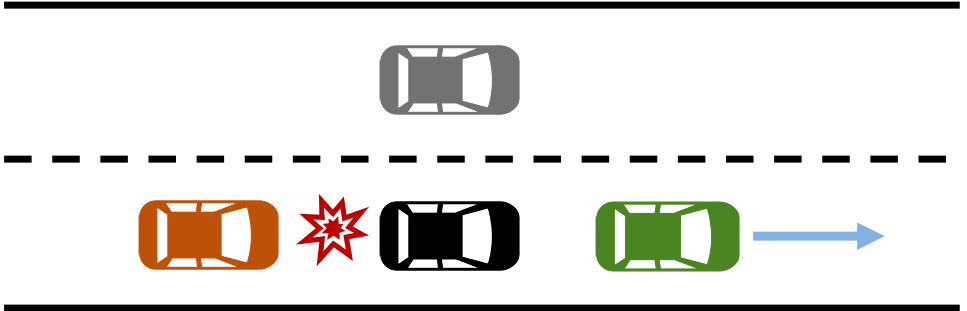}
    \caption{Rear-end collision}
    \begin{minipage}{3cm}
    \vfill
    \end{minipage}
  \end{subfigure}
  \caption[Motivational scenarios]{Motivational scenarios. AV is in orange. (a) The AV is struck by an oncoming vehicle going straight (in green) when making a left turn, as its view is obstructed by the truck. (b) The AV crashes into the back of the vehicle in front of it as the leading vehicle (in green) brakes abruptly.} \label{figchap4:motivation}
\end{figure}

Predicting the trajectories of surrounding vehicles is a critical component of ensuring collision-free path planning. Autonomous vehicles (AVs) provide promising solutions to achieving safer road environment \cite{fagnant2015preparing} facilitated by the technological advancement in sensors (camera, LiDAR, radar, etc) and artificial intelligence. Connected vehicle (CVs) technologies are also being deployed to improve the road safety by enhancing vehicular situational awareness through various CV options such as vehicle-to-vehicle (V2V) and vehicle-to-infrastructure (V2I) communications. With the same objective of improving road safety, CVs and AVs really should be developed in a cooperative manner. As argued in \cite{shladover2018connected}, CV technology provides information that AV sensors can not see immediately (e.g. maneuver commands issued to other vehicles) as well as information that may be occluded by adjacent vehicles. Two examples are shown in Figure~\ref{figchap4:motivation}. In both scenarios, the AV failed to detect the potential hazard in advance due to its limited field of view. However, if the green vehicle could communicate its trajectory information to the AV, the accidents could be effectively avoided. We believe that the integration of AV and CV technologies can provide complementary benefits, significantly enhancing the performance and safety of the transportation system. Hereinafter, we refer to a vehicle with attributes from both sides as connected and autonomous vehicle (CAV), whereas vehicles without these attributes are termed Human-Driven Vehicles (HDVs). 

While extensive research has focused on trajectory prediction within the context of AV, trajectory prediction studies within the CAV environment have garnered comparatively less attention, primarily due to the low market penetration rate (MPR) of CVs. Gathering real-world data in the CAV environment is inherently more complex than in purely autonomous settings. Given that CAVs are still in the early stages of deployment, obtaining large-scale datasets with comprehensive trajectory information from actual CAVs navigating intricate traffic scenarios remains a limitation. Most CAV trajectory prediction studies rely on datasets like the Next Generation Simulation (NGSIM), which, regrettably, lacks CAV-specific data. Consequently, it may not fully encapsulate the distinct characteristics and behaviors exhibited by CAVs. 

Before the full-scale deployment of either connected or autonomous vehicular technology, it is anticipated that a prolonged transition period will ensue, characterized by a mixed traffic environment consisting of CVs, AVs, CAVs and HDVs. With the gradual deployment of CAV technology in real-world setting, studying the trajectory prediction in a CAV environment can help identify the limitations and robustness for trajectory prediction models that need to operate in complex and dynamic traffic environment. 

\begin{figure}[!ht]
  \centering
  \includegraphics[width=0.7\textwidth]{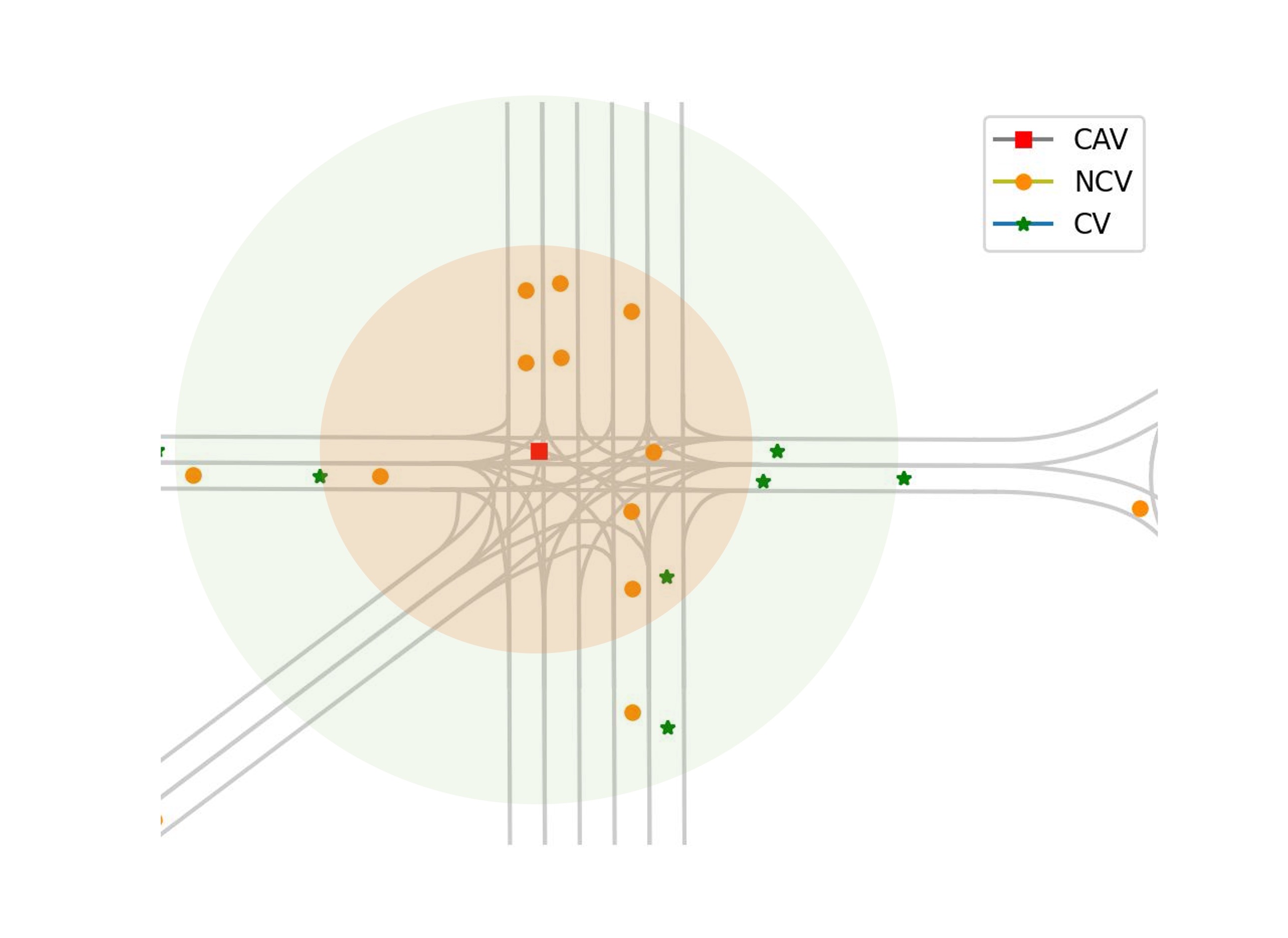}
  \caption[Agents in a mixed traffic flow]{The mixed traffic flow consists of a CAV as central agent (red square), CVs (green stars), AVs, and HDVs. AVs and HDVs are grouped into NCVs (orange dots). The CAV learn about their surroundings through sensors and communication technology. The communication range (green circle) is typically greater than the sensing range (orange circle). Please note: if there are other CAVs in the surroundings of the central CAV, it will be considered as a CV, as their trajectories are broadcasted to the central CAV through communication.}
  \label{figchap4:scenario}
\end{figure}

 To address this gap, we investigate a mixed traffic scenario in which a CAV serves as the central agent, as illustrated in  Figure~\ref{figchap4:scenario}. The CAV collects information from its surroundings using both sensors and communication technology. Typically, the communication range extends farther than the sensing range, covering approximately 300 meters for the former and 200 meters for the latter \cite{vargas2021overview}. As a result, connected vehicles within the sensing zone provide dual observations. It is worth noting that we categorize both AVs and HDVs as non-connected vehicles (NCVs) since information regarding these two categories is solely acquired through sensors. It is crucial to recognize that data obtained from different sources inherently carry distinct types of errors, with sensor noise and communication latency being the most common ones. Leveraging the complementary nature of information from these sources holds the potential to enhance scene understanding and, importantly, mitigate errors associated with each individual source. To facilitate our study, we collected a trajectory dataset in the CARLA simulator \cite{dosovitskiy2017carla}, introducing synthesized data errors tailored to each source to accurately reflect their inherent characteristics. 
 
 A novel deep learning framework was proposed comprising a multi-source encoder, an agent-agent and agent-lane interaction module and a multi-agent decoder. The multi-source encoder efficiently processes historical vehicle trajectories by employing source-specific temporal encoders to capture the temporal dependencies and a cross-attention fusion module for multi-source data integration. By representing each agent and lane as graph nodes, we leverage Graph attention models (GAT) to capture agent-agent and agent-lane interactions. The multi-agent decoder simultaneously predicts the trajectories of all surrounding vehicles. An Ablation study is conducted to assess the impact of multi-source data and the cross-attention fusion module on the model’s performance. Furthermore, we compare model's performance across different MPRs to unveil the synergistic benefits of sensor and communication technologies. Additionally, we illustrate the prediction results on individual scenes. Finally, a comparative study with existing research work is presented. 

The main contributions of this work lie in three-folds:
\begin{itemize}
  \item We are the first to study the impact of multi-source data for trakectory prediction in a mixed traffic flow.
  \item A customized motion dataset with map geometry was collected in CARLA and utilized for the validation of the problem. 
  \item A novel deep learning framework is proposed to handle the multi-source data fusion, agent-agent and agent-lane interactions, and multi-agent predictions.
\end{itemize}

\section{Related Work}
\subsection{Publicly Available Motion Dataset}
There is a plethora of open motion datasets collected by autonomous vehicle on-board sensors in urban scenarios such as Waymo  \cite{ettinger2021large}, ArgoVerse \cite{chang2019argoverse}, ArgoVerse2 \cite{wilson2021argoverse}, NuScenes \cite{caesar2020nuscenes}, etc., which have significantly advanced trajectory prediction research within the context of AVs. Research that utilizes these datasets often assumes a flawless data source and does not account for the inherent data errors they may encompass. The NGSIM dataset, commonly employed in the context of CAV, was collected by video cameras mounted on buildings and lacks CAV specific scenarios. Primarily, it encompasses highway driving scenarios. The V2X-Seq trajectory dataset \cite{yu2023v2x} provides valuable multi-view trajectory information, which is closely related to our study. However, our work differs as we focus on the overlapping range of on-board sensors and communication technologies, where complementary information can be utilized. To the best of our knowledge, there are currently no publicly available motion datasets dedicated to CAV research. Therefore, we will leverage a simulated dataset generated by CARLA, affording us the flexibility to construct CAV scenarios and explore data sources featuring diverse characteristics.

\subsection{Trajectory Prediction in Connected and Autonomous Vehicle Environment}
\subsubsection {Autonomous Environment}
Due to the temporal-spatial characteristics of trajectory data, recurrent neural networks (RNN) with long short-term memory (LSTM) or gated recurrent unit (GRU) architectures are extensively employed to predict future trajectories taking their historic trajectory as input  \cite{altche2017lstm,kim2017probabilistic,park2018sequence}. Social-LSTM was proposed considering interactions among neighboring vehicles \cite{deo2018would, lin2021vehicle}. The spatial information was encoded by convolution neural networks (CNN) or graph neural networks (GNN) \cite{li2019grip++}. As vehicles' decisions are largely associated with the map topology, an emerging avenue of research is proposed to leverage the high-definition maps (HD maps) to assist the prediction. \cite{cui2019multimodal, chai2019multipath, djuric2020uncertainty} rasterized the traffic scenes into birds' eye view images where map vector layers (e.g. polygons, polylines) are encoded into RGB color space. Conventional CNN are then used to model the scene. As the image-based representation is dense and computationally expensive, subsequent work proposed to use more compact and efficient vectorized representation (polygons, polylines) directly \cite{zhao2020tnt,gao2020vectornet,khandelwal2020if,varadarajan2022multipath++} and then model by graph convolutional networks (GCN) \cite{liang2020learning}, transformers \cite{zhou2022hivt}, or point cloud networks \cite{ye2021tpcn}. 

\subsubsection{Connected Environment}
Trajectory prediction models proposed in CV lags behind the ones developed in AV as autonomous driving has a much higher requirement on the trajectory prediction accuracy. \cite{goli2018vehicle} first clustered the training data then employed Gaussian process regression to learn the motion pattern. \cite{xing2019personalized} recognized three driving styles using Gaussian mixture model, then proposed a personalized joint time series model based on LSTM to predict the front vehicle trajectories. \cite{mo2020interaction} proposed a CNN-LSTM framework utilizing 8 surrounding vehicles to predict the future trajectory of an ego-vehicle. \cite{lu2022vehicle} developed a heterogeneous context-aware graph convolutional networks which combines the information encoding of historical vehicle trajectories, scene image stream and inter-vehicle interaction patterns to predict the future trajectory of a target vehicle. 

\subsubsection{Connected and Autonomous Environment}
There is limited literature in the CAV context. \cite{lin2021long} proposed to utilize all the historic trajectories of surrounding CAVs to predict the longitudinal trajectory of a HDV in front of the ego-vehicle. Promising results showed that information from CAV improved the model performance even when the market penetration rate is as low as 20\%. \cite{lv2022trajectory} proposed an interactive network model utilizing LSTM and CNN with a correction mechanism. The model enables a CAV to predict the trajectories of surrounding vehicles, and the trajectories can be corrected during the prediction process to improve the prediction accuracy. Both models are tested on the NGSIM dataset which contains only limited driving scenarios: highway driving (including ramp merging and double lane change) and signalized intersection scenarios. 

We will essentially follow a similar approach for modeling temporal, spatial, and contextual information, with the distinction of operating within a CAV environment where we will account for multi-source data integration.

\subsection{Multi-source Data Fusion in Cooperative Driving}
There is a growing trend in leveraging multi-source data through Vehicle-to-Everything (V2X) communications to enhance model performance. However, existing research primarily focuses on cooperative perception, where perceptions from different viewpoints are fused \cite{xu2022v2x, cui2022coopernaut}. Some studies have considered joint perception and motion forecasting, but the emphasis on data fusion persists mainly in the perception stage \cite{wang2020v2vnet}. With the recent release of a V2X trajectory dataset \cite{yu2023v2x}, we anticipate an increase in research directly addressing trajectory data fusion.

\section{Data Preparation and Scene Representation}
\subsection{Data Preparation}

\begin{figure}[h!]
    \centering
    \subfloat[\centering CARLA simulation]{{\includegraphics[width=.45\textwidth, height=6cm]{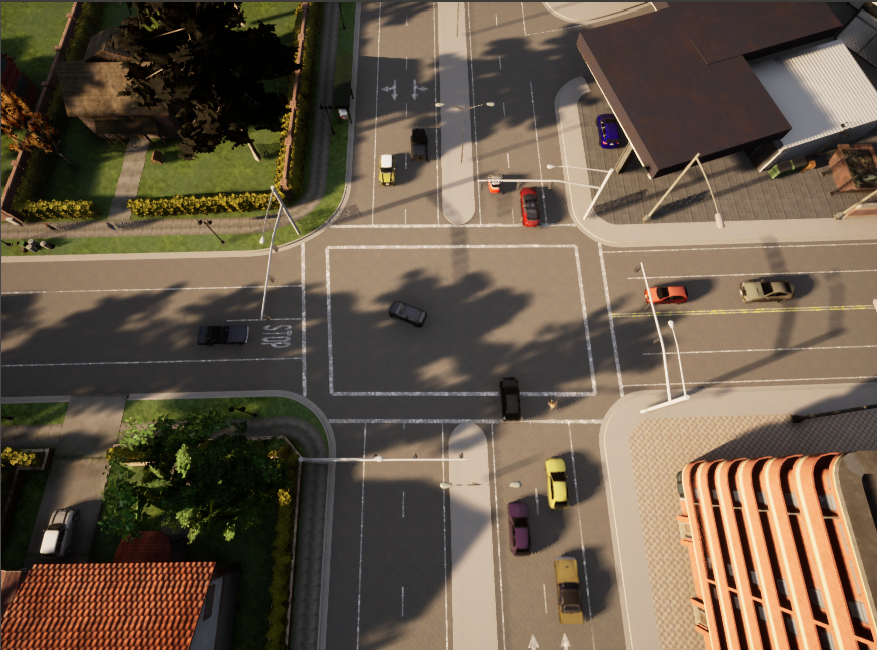} }}%
    \hspace{-1cm}
    \subfloat[\centering Town03 map]{{\includegraphics[width=.5\textwidth, height=6cm]{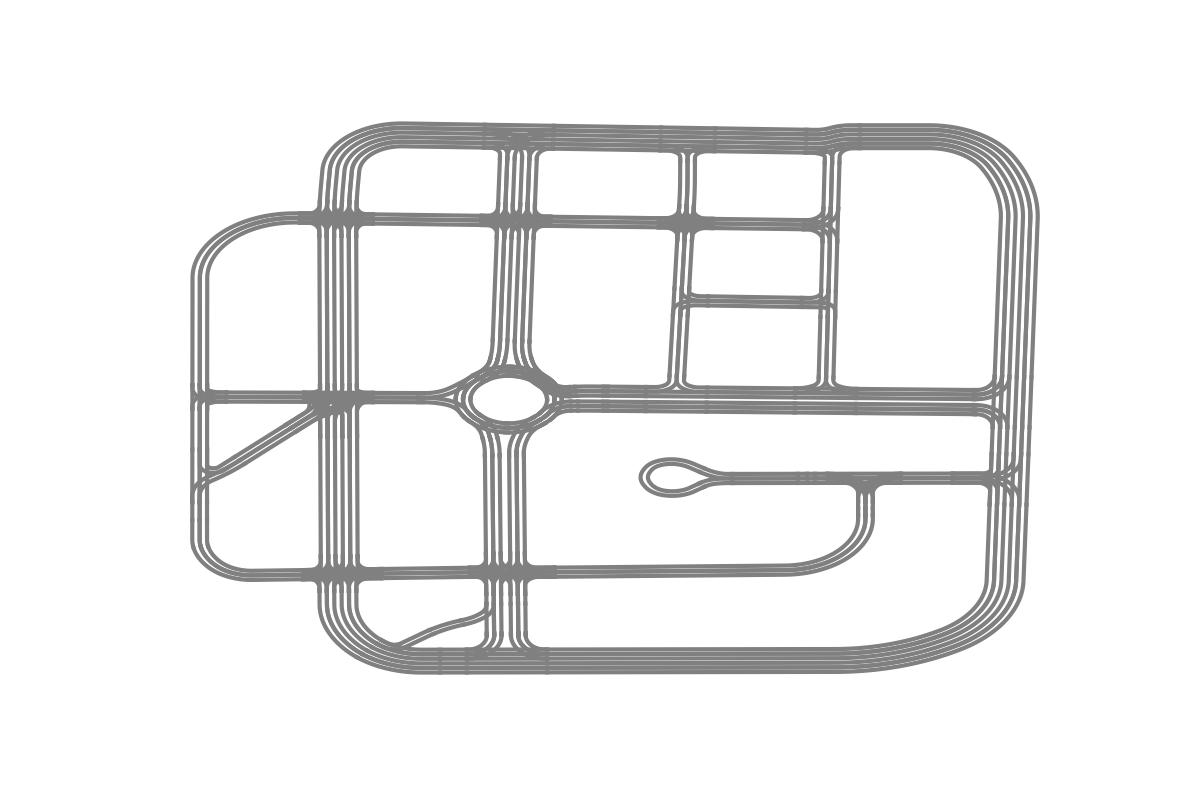} }}%
    \caption[CARLA simulator]{(a) Traffic simulation in CARLA simulator. (b) Map of CARLA Town03}%
    \label{figchap4:carla}%
\end{figure}

The traffic was simulated in CARLA Town03 using the Traffic Manager (TM) module, which controls vehicles in autopilot mode. The simulation interface is shown in Figure~\ref{figchap4:carla}(a). The simulation lasted for 270 seconds with a sampling rate of 10Hz. The positions of the 265 populated vehicles were logged at each timestamp. 

To construct the scenario depicted in Figure~\ref{figchap4:scenario}, we employed the following process on the collected dataset: First, we iterated through all the vehicles designated as CAVs, one at a time. For each CAV at time t, we identified all the neighboring vehicle IDs within a 50-meter radius which is the communication range we employed. Among these IDs, we randomly assigned CV IDs according to the MPR. We also identified all the IDs that fall within a 30-meter sensing range. The trajectory data for these identified IDs were processed within a time window spanning from t-5 seconds to t+5 seconds, constituting a 10s time window. Within this window, we allocated 3s for motion history and 5s for the prediction horizon. The initial 2s were reserved for simulating latency errors. 

To simulate latency, we introduce delays ranging from 1 to 15 frames to the historical trajectories of CV IDs, shifting the ground-truth trajectories. Additionally, Gaussian noises with variances ranging from 0 to 0.5 are introduced to the historical trajectories of sensor IDs. It's important to note that for IDs found in both CV IDs and sensor IDs, we maintain dual sets of historical trajectories — one representing the delayed data and the other the noisy data. This approach allows us to account for both latency and sensor noise in the dataset.

The map of Town03 is publicly available in OpenStreetMap (OSM) XML format (shown in Figure~\ref{figchap4:carla}(b)) and is processed following the Argoverse map API (Chang, 2019). We utilize the lane centerlines within a 75-meter radius of the CAV. 

\subsection{Scene Representation}
Vectorized representation was employed for its efficiency \cite{gao2020vectornet}. The scenes were first transformed such that the CAVs are positioned at the origin with its heading aligned along the positive x-axis.
Specifically, each trajectory can be represented as a sequence of displacements  $\{{\Delta \mathbf{p}_{t}\}}^0_{t=-(T_h-1)}$, where $\Delta \mathbf{p}_{t} \in \mathbb{R}^2$ is the 2D displacement from time step $t-1$ to $t$ and $T_h$ is the total historical time steps. Similarly, for a lane centerline composed of lane segments, each segment can be represented as $\Delta \mathbf{p}_l^k \in \mathbb{R}^2$, which captures the 2D displacement from the starting coordinate to the end coordinate of lane segment $k$.

\section{Methodology}
\subsection{Overview}
\begin{figure}[!ht]
  \centering
  \includegraphics[width=1\textwidth]{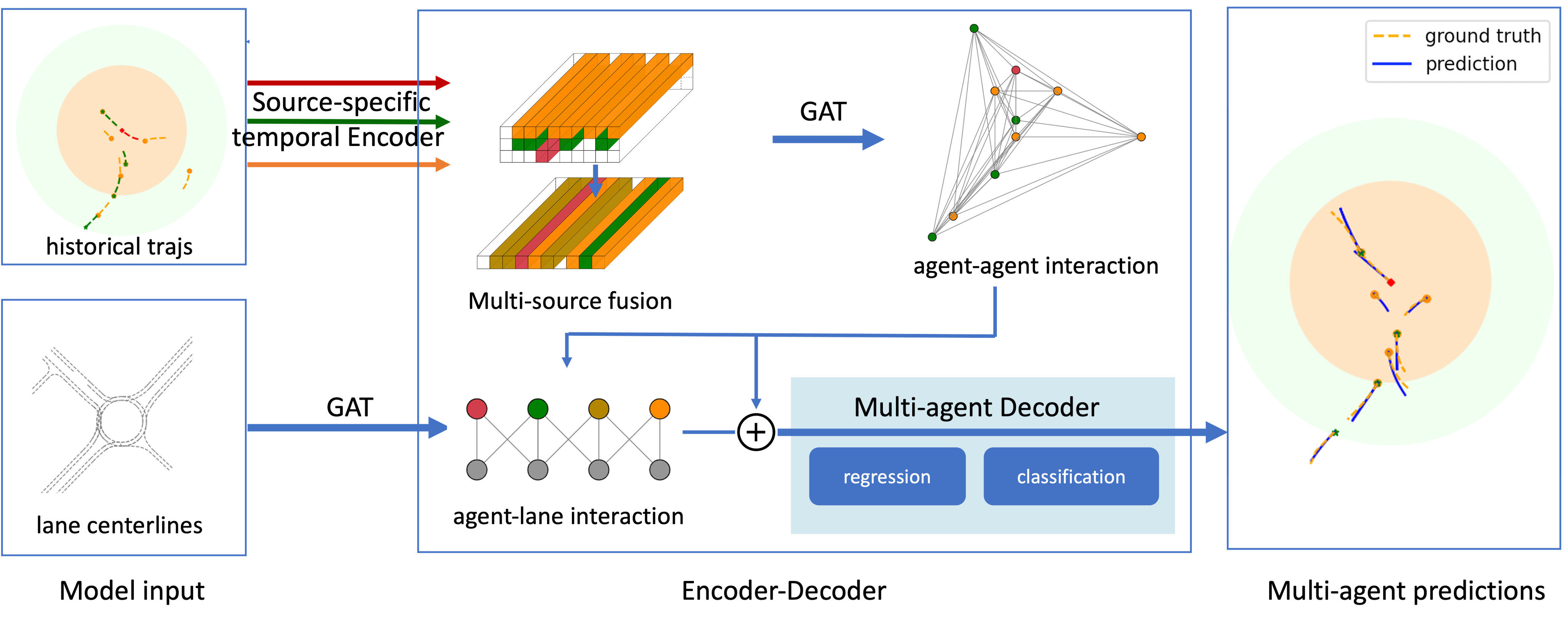}
  \caption[Proposed MSMA model architecture]{Proposed model architecture. Synthesized source-specific errors are introduced to the historical trajectories and then each source is encoded by a source specific temporal encoder. Cross-attention is employed to effectively fuse data from dual sources. Each agent is then modeled as a graph node with the temporal feature being its node feature. GAT is leveraged to capture the agent-agent and agent-lane interaction. The resulting encodings from both GATs are concatenated as input to a multi-agent decoder for predicting future trajectories of all target agents.}\label{figchap4:arch}
\end{figure}

An overview of our proposed model is illustrated in Figure~\ref{figchap4:arch}. It consists of a multi-source temporal encoder, a Graph Attention Network (GAT) based agent-agent encoder, an agent-lane encoder and a multi-agent decoder. 

\subsection{Multi-source Temporal Encoder}
To differentiate between data obtained from sensors and communication technologies, we employ source-specific temporal encoders for capturing temporal dependencies. While both RNNs and Transformers are suitable for handling temporal data, we use Transformer as the base model due to its efficiency in capturing long-range dependencies. Specifically, a self-attention mechanism is employed which allows the agent to focus on relevant temporal information from its own historical trajectory. 

We denote the input from agent $i$ and source $s$ as $\{\Delta \mathbf{p}_{t}^{i_s}\}^0_{t=-(T_h-1)}$, where $s \in \{1,2,3\}$. Source 1 represents the input directly from CAV itself, for which we utilize ground-truth data in this context. Source 2 corresponds to input from communication technologies, where latency has been intentionally introduced. Source 3 captures input from sensor technologies which contains noise. 

A multilayer perceptron (MLP) is applied to the input to get its embedding $\mathbf{Z}^{i_s} \in \mathbb{R}^{T_h \times d_h}$, where $d_h$ is the embedding dimension. Following BERT \cite{devlin2018bert} and HiVT \cite{zhou2022hivt}, we then append a learnable token $\mathbf{z}^{i_s} \in \mathbb{R}^{d_h}$ to the end of the embedding. Learnable positional encodings are then added to the embedding. The updated embedding $\mathbf{Z}^{i_s} \in \mathbb{R}^{(T_h+1) \times d_h}$ is then fed into the self-attention block:
\begin{linenomath}
    \begin{flalign} 
        &\mathbf{Q}^{i_s}=\mathbf{Z}^{i_s}\mathbf{W}^{Q_s}, \mathbf{K}^{i_s}=\mathbf{Z}^{i_s}\mathbf{W}^{K_s}, \mathbf{V}^{i_s}=\mathbf{Z}^{i_s}\mathbf{W}^{V_s}\\
        &\tilde{\mathbf{Z}}^{i_s} = \frac{softmax(\mathbf{Q}^{i_s}\mathbf{K}^{i_sT})}{\sqrt{d_k}}\mathbf{V}^{i_s}
    \end{flalign}
\end{linenomath}
where $\mathbf{W}^{Q_s}$, $\mathbf{W}^{K_s}$, $\mathbf{W}^{V_s} \in \mathbb{R}^{d_h \times d_k}$ are learnable matrices for linear projection and $d_k$ is the dimension of the transformed vectors. $1/\sqrt{d_k}$ implements the scaled dot product term for numerical stability of attention. Multiple heads can be used to attend to different parts of the sequence simultaneously. In addition, we apply Dropout and Layer Normalization to stabilize the training process. The updated appended token $\tilde{\mathbf{z}}^{i_s} \in \mathbb{R}^{d_k}$ summarizes of the temporal features. For agents with a single data source, this is the input to the subsequent agent-agent interaction module.

For agents with dual data sources, we apply across-attention to align and fuse the temporal features across the two sources. The temporal feature from sensor source is converted to the query vector and the temporal feature from communication source is used to compute the key and value vectors:

\begin{linenomath}
    \begin{flalign} 
        &\mathbf{\tilde{q}}^{i_3}=\mathbf{\tilde{z}}^{i_3}\mathbf{W}^{Q_f}, \mathbf{\tilde{k}}^{i_2}=\mathbf{\tilde{z}}^{i_2}\mathbf{W}^{K_f}, \mathbf{\tilde{v}}^{i_2}=\mathbf{\tilde{z}}^{i_2}\mathbf{W}^{V_f}\\
        &\mathbf{F}^{i_3} = \frac{softmax(\mathbf{\tilde{q}}^{i_3}\mathbf{\tilde{k}}^{i_2T})}{\sqrt{d_k}}\mathbf{\tilde{v}}^{i_2}\\
        &\alpha^{i} = sigmoid(\mathbf{W}^{\alpha}[\mathbf{\tilde{z}}^{i_3}, \mathbf{F}^{i_3}])\\
        &\mathbf{\hat{Z}}^{i} = \alpha^{i} \odot \mathbf{W}^{self}\mathbf{\tilde{z}}^{i_3}+(1-\alpha^{i})\odot\mathbf{F}^{i_3}
    \end{flalign}
\end{linenomath}

where $\mathbf{W}^{Q_f}$, $\mathbf{W}^{K_f}$, $ \mathbf{W}^{V_f}$, $\mathbf{W}^\alpha$, $\mathbf{W}^{self}\in R^{d_k \times d_k}$ are learnable matrices and $\odot$ denotes element-wise product. $\mathbf{F}^{i_3}$ represents the aligned features from the communication source in the sensor source. We then apply sigmoid to obtain a weight vector which computes the final fused representation ${\hat{\mathbf{z}}}^i\in R^{d_k}$ from the two sources. ${\hat{\mathbf{z}}}^i$ is then the input for agents with dual sources to the next module as illustrated in Figure~\ref{figchap4:arch}.

\subsection{Agent-agent Interaction}
To capture agent-agent interactions, we employ the GAT architecture \cite{velivckovic2017graph} to attend to neighboring features. Each agent is represented as a node within the graph. The temporal features ${\tilde{\mathbf{z}}}^i$ (${\hat{\mathbf{z}}}^i$ for agents with dual source) we have just obtained serve as node features. The relative positions $\mathbf{p}^i_0-\mathbf{p}^j_0$ between two agents $i$ and $j$ at the last observed step $t=0$ are modeled as edge attributes. GAT dynamically adjusts the weight of each edge during the message-passing process. We utilize multiple stacked graph attention layers to compute and update the edge weight $e_{ij}$. The resulting weighted information aggregation from surrounding agents of agent $i$ can be expressed as follows:
\begin{linenomath}
    \begin{flalign}
    &e_{ij} = a_G(\Tilde{\mathbf{z}}^i\mathbf{W}^G, \Tilde{\mathbf{z}}^j\mathbf{W}^G, (\mathbf{p}^i_0-\mathbf{p}^j_0)\mathbf{W}^G) \label{eqchap4:11}\\
    &\alpha_{ij} = Softmax(e_{ij}) \\
    &\mathbf{S}^l(i) = \sum_{j\in N_i} \alpha_{ij}\Tilde{\mathbf{z}}^j\mathbf{W}^G
    \end{flalign}
\end{linenomath}
where $W^G \in \mathbb{R}^{d_k \times d_g}$ is the shared linear transformation, $d_g$ is the new feature dimension, the shared attentional mechanism $a_G$ is a single-layer feedforward neural network: $\mathbb{R}^{d_k} \xrightarrow{} \mathbb{R}$ that computes the edge weight $e_{ij}$ indicating the importance of vehicle $j$'s feature to vehicle $i$, and $S^l$ is the output of the $l$th layer by summing up the weighted features from all the neighboring vehicles. Output from the last layer $\tilde{l}$ is the updated node feature, denoting as $\tilde{\mathbf{S}}^i$. 

\subsection{Agent-lane Interaction}
Now we consider the context information provided by lane centerlines. Each lane segment $\Delta \mathbf{p}_{l}^k \in \mathbb{R}^2$ is represented as a graph node, where the node feature is obtained by an MLP embedding $\mathbb{R}^2 \xrightarrow{} \mathbb{R}^{d_g}$:
\begin{linenomath}
    \begin{equation} 
        \mathbf{L} = MLP_{L}(\Delta \mathbf{p}_{l}^k)
    \end{equation}
\end{linenomath}
We have $\epsilon$ total segments queried within a 75m radius of the CAV. To efficiently assess relevant lane information while mitigating the computational load associated with distant or non-essential segments, we employ a 30m local radius for each traffic agent. Consequently, only lane segments falling within this local radius will be taken into account when evaluating interactions.

The edge attribute between a lane segment $k$ and an agent $i$ is determined by their relative position $\mathbf{p}_l^k-\mathbf{p}^i_0$. If lane node $k$ and agent node $i$ are connected, we calculate the edge weight $e_{ik}$ using graph attention layers:
\begin{linenomath}
    \begin{flalign} 
        &e_{ik} = a_{L}(\mathbf{L}^k\mathbf{W}^L,  \Tilde{\mathbf{S}}^i\mathbf{W}^L, (\mathbf{p}_l^k-\mathbf{p}^i_0)\mathbf{W}^L)\\
        &\alpha_{ik} = Softmax(e_{ik}) \\
        &\mathbf{M}^l(i) = \sum_{k\in N_i} \alpha_{ik}\mathbf{L}^k\mathbf{W}^L
    \end{flalign}
\end{linenomath}
where $\mathbf{W}^L \in \mathbb{R}^{d_g \times d_g}$ is the shared linear transformation, $a_L$ is a feedforward neural network as in Eq. \ref{eqchap4:11} and $\mathbf{M}^l$ represents the aggregated lane information from the $l$-th layer. We output the node features from the last layer $\tilde{l}$ node feature, denoting as $\tilde{\mathbf{M}}^i$, which is then concatenated with $\tilde{\mathbf{S}}^i$ as the final temporal-spatial embedding for agent $i$. 

\subsection{Multi-agent Decoder}
We parameterize the distribution of future trajectories as a mixture model where each mixture component is a Gaussian distribution. The predictions are made for all agents in a single shot. For each agent, the model predicts $D$ possible future trajectories (mixture components) and their confidence scores (mixing coefficients). The decoder has two headers: a regression head to predict the trajectory of each mode and a classification head to predict the confidence score of each mode. For agent $i$ and its mode $d$, in the regression head, an MLP receives the final embedding as inputs and outputs the location $\mu^t_{i,d} \in \mathbb{R}^2$ and its associated covariance ${\sigma^2}^t_{i,d} \in \mathbb{R}^{2*2}$ of the agent per future time step. In the classification head, we use another MLP followed by a softmax function to produce the confidence scores for each mode.
\subsection{Model Training}
Since all the modules are differentiable, we can train the model end-to-end. The Loss function is defined as
\begin{linenomath}
    \begin{equation}
      J = J_{reg} + \alpha J_{cls} 
    \end{equation}
\end{linenomath}
Here, $J_{reg}$ represents the loss from the regression head, and $J_{cls}$ represents the classification loss. After experimentation, we have set $\alpha$ to 0.5. Given a set of $D$ predicted trajectories, we identify the trajectory $\tilde{d}$ with the minimum Average Displacement Error (ADE), which is defined as the average Euclidean distance between ground-truth locations and predicted locations across all future time steps. We utilize the negative log-likelihood to calculate the regression loss:
\begin{linenomath}
    \begin{equation}
      J_{reg} = -\frac{1}{nT_f}\sum^n_{i=1} \sum^{T_f}_{t=1} logP(\mathbf{p}_t^{i,pred} - \mathbf{p}_t^{i,gt} | {\hat{\mu}}^{i,\tilde{d}}_t,{\hat{\sigma^2}}^{i,\tilde{d}}_t)
    \end{equation}
\end{linenomath}
where $T_f$ is the prediction horizon, n is the number of agents, $P(|)$ is the probability density function of Gaussian distribution and $\hat{\mu}_t^{i,\tilde{d}}$, $\hat{\sigma^2}_t^{i,\tilde{d}}$ are the mean and covariance from predicted trajectory $\tilde{d}$.

For $J_{cls}$, cross-entropy loss is applied:
\begin{linenomath}
    \begin{equation}
      J_{cls} = -\frac{1}{n}\sum^n_{i=1} \sum^{D}_{d=1} \mathbbm{1}_d log{p_d}
    \end{equation}
\end{linenomath}
where $\mathbbm{1}_d$ is a binary indicator if $d$ is the best mode, $p_d$ is the confidence score of mode $d$.

The model was trained for 100 epochs using AdamW optimizer with initial learning rate 1e-3 for 10 warm-up epochs. The learning rate was then decayed every 10 epochs by a factor of 0.9. Our model consists of 4 layers for temporal encoding, 8 layers for both agent-agent and agent-lane interactions, 8 heads for all attention modules. We used 16-dimensional input embedding size and 64 for all hidden dimensions. The dropout rate was 0.1 throughout the architecture.
\section{Results Analysis}
The complete dataset was split into train, validation and test set with a ratio of 8:1:1. The standard metrics in motion predictions are adopted, including Average Displacement Error (ADE), Final Displacement Error (FDE), and Miss Rate (MR), where errors between the best predicted trajectory among the D=5 modes and the ground truth trajectory are calculated. The best here refers to the trajectory that has the minimum endpoint error. The ADE metric calculates the L2 distance across all future time steps and averages over all scored vehicles within a scenario, while FDE measures the L2 distance only at the final future time step and summarizes across all scored vehicles. MR refers to the ratio of actors in a scenario where FDE are above 2 meters.
\subsection{Ablation Study}
To demonstrate the impact of cross-attention fusion, which aligns input from sensor and communication sources, we conducted two ablation studies under MPR=0.8. 

\begin{figure}
  \begin{subfigure}{0.48\textwidth}
    \centering
    \includegraphics[width=\textwidth]{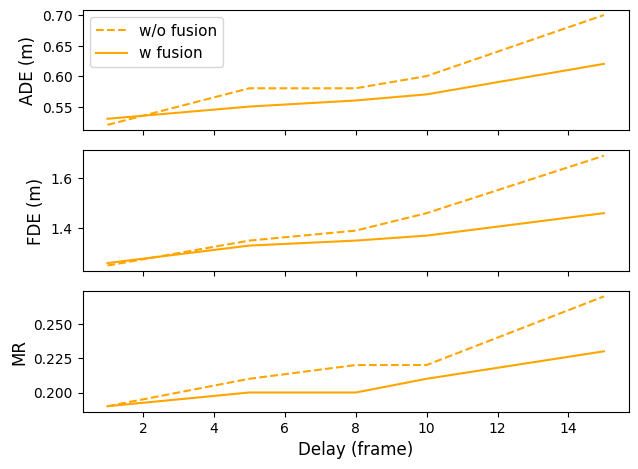} 
    \caption{}
  \end{subfigure}
  \begin{subfigure}{0.48\textwidth}
    \centering
    \includegraphics[width=\textwidth]{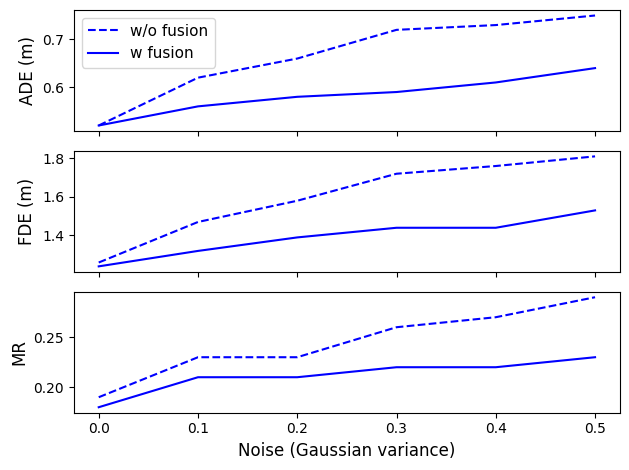}
    \caption{}
  \end{subfigure}
  \caption[Ablation study for fusion module]{Ablation study for fusion module. (a) Prediction performance for all the CVs in a scene when different time delays are introduced into their historical trajectories. (b) Prediction performance for all the vehicles within sensing range in a scene when Gaussian noises with different variance are introduced into their historical trajectories. } \label{figchap4:ablation}
\end{figure}

In the first study, our objective was to assess the fusion module's resilience to communication latency. This involved introducing a time delay to all historical CV trajectories, varying from 1 to 15 frames, i.e., 0.1 to 1.5 seconds. Gaussian noise was held constant at 0 for sensor trajectories. We then compared the ADE, FDE and MR metrics for all CVs under the model architectures with and without the cross-attention fusion module. The results, presented in Figure~\ref{figchap4:ablation}(a), clearly indicate that the inclusion of fused information from error-free sensors enhances the robustness of CVs against communication latency.

In the second study, we aim to evaluate the fusion module's ability to withstand sensor noise. We introduced Gaussian noises with varying variances, ranging from 0 to 0.5, to all historical trajectories of vehicles within the sensing range, while no latency was introduced to CV trajectories in this evaluation. The metrics comparison is shown in Figure~\ref{figchap4:ablation}(b), It is evident that the incorporation of fused information from communication technologies effectively mitigated the impact of sensor noises.
\subsection{Performance under Different CV Market Penetration Rates}
\begin{figure}
  \centering
  \includegraphics[width=.6\textwidth]{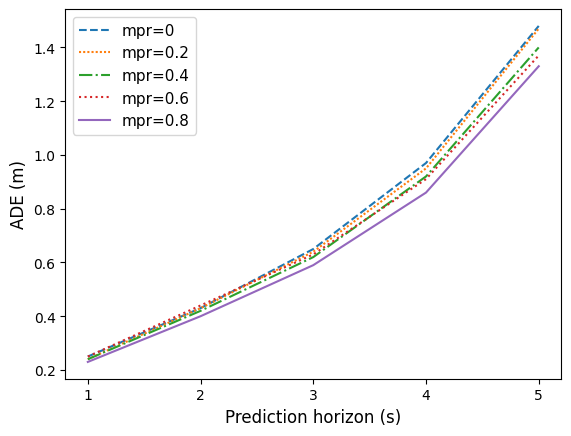}
  \caption[Model performance comparison under different CV MPRs]{Model performance comparison on vehicles within sensing range under different CV MPRs. Higher MPRs lead to lower ADE values.}\label{figchap4:mpr}
\end{figure}

We are also interested in assessing how varying numbers of CVs in the scenario influence the model's performance. To conduct this evaluation, models are trained under different MPRs with a fixed 1-frame delay and 0.1 noise variance. Since scenarios corresponding to different MPRs consist of varying numbers of CVs, to ensure a fair comparison, we evaluate the prediction performance using ADE exclusively for vehicles within the sensor range across different MPRs. The evaluation results are presented in Figure~\ref{figchap4:mpr}. We can observe a clear trend of performance improvement as the MPR increases especially at longer prediction horizons. 

\begin{table}[h!]
\caption[Performance evaluation under different CV MPRs]{Performance evaluation under different CV MPRs. \label{tabchap4:mprcompr}}
\begin{center}
\resizebox{0.7\textwidth}{!}{%
\begin{tabular}{c|ccccc}
\hline 
Metrics&MPR=0&MPR=0.2&MPR=0.4&MPR=0.6&MPR=0.8 \\
\hline
ADE&0.62&0.61&0.59&0.59&0.56\\

FED&1.48&1.47&1.40&1.37&1.33\\

MR&0.23&0.22&0.22&0.21&0.20\\
\hline
\end{tabular}%
}
\end{center}
\end{table}

\begin{figure}
  \centering
  \begin{subfigure}{0.4\textwidth}
    \includegraphics[width=\linewidth]{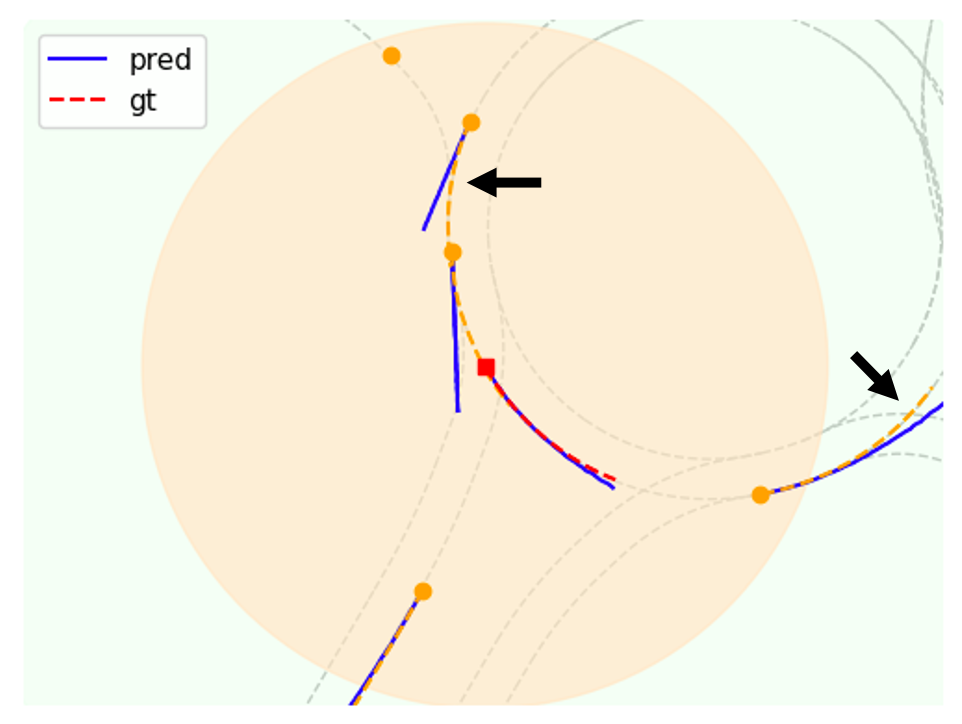}
    \caption{MPR=0, ADE=0.74}
  \end{subfigure}
  \begin{subfigure}{0.4\textwidth}
    \includegraphics[width=\linewidth]{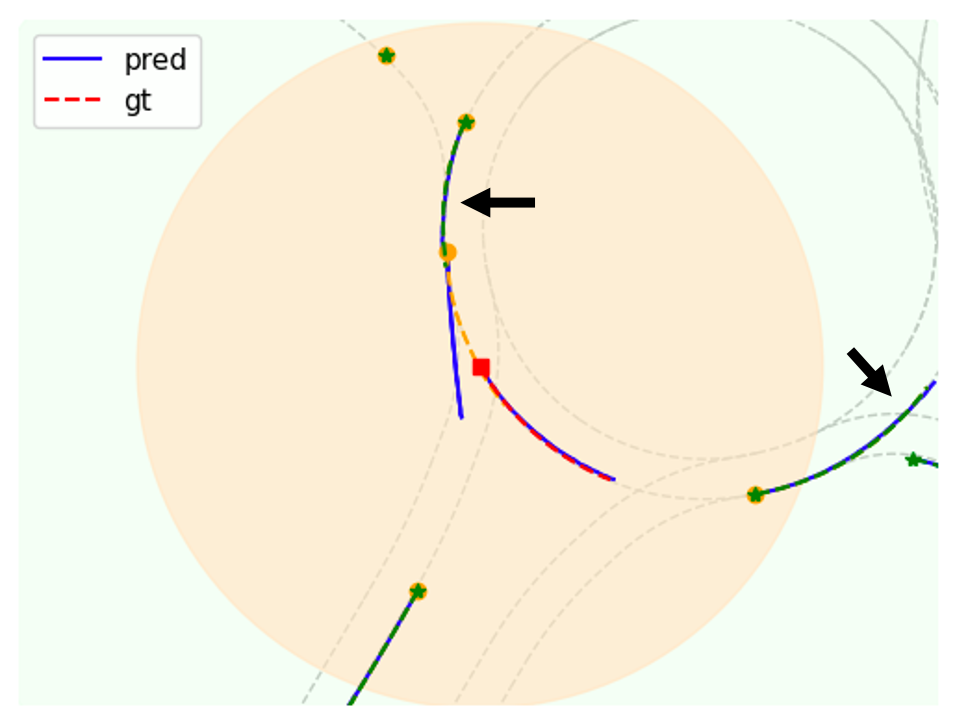}
    \caption{MPR=0.8, ADE=0.48}
  \end{subfigure}
  
  \caption[Visualization of multi-agent prediction in a roundabout]{Visualization of multi-agent prediction in a roundabout. Lane centerlines are shown in the background. ADE for vehicles within the sensing range (orange filled circle) are computed for performance comparison. Dashed line represents ground truth while solid line represents prediction. Prediction enhancements are indicated by black arrows.
  }\label{figchap4:s480_mpr}
\end{figure}

\begin{figure}
  \centering
  \begin{subfigure}{0.4\textwidth}
    \includegraphics[width=\linewidth]{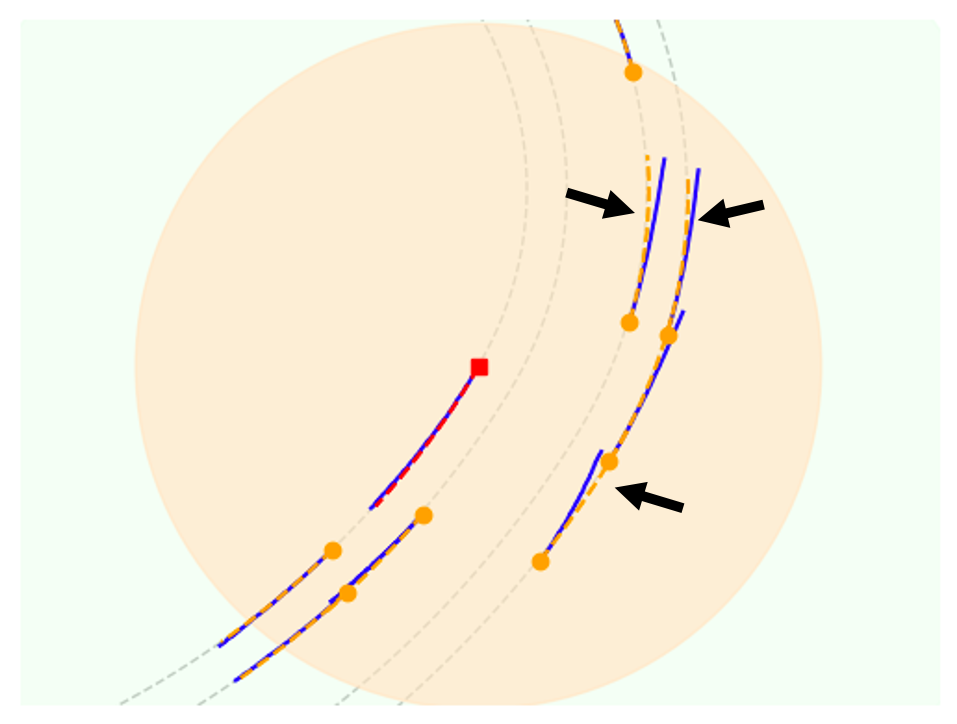}
    \caption{MPR=0, ADE=0.69}
  \end{subfigure}
  \begin{subfigure}{0.4\textwidth}
    \includegraphics[width=\linewidth]{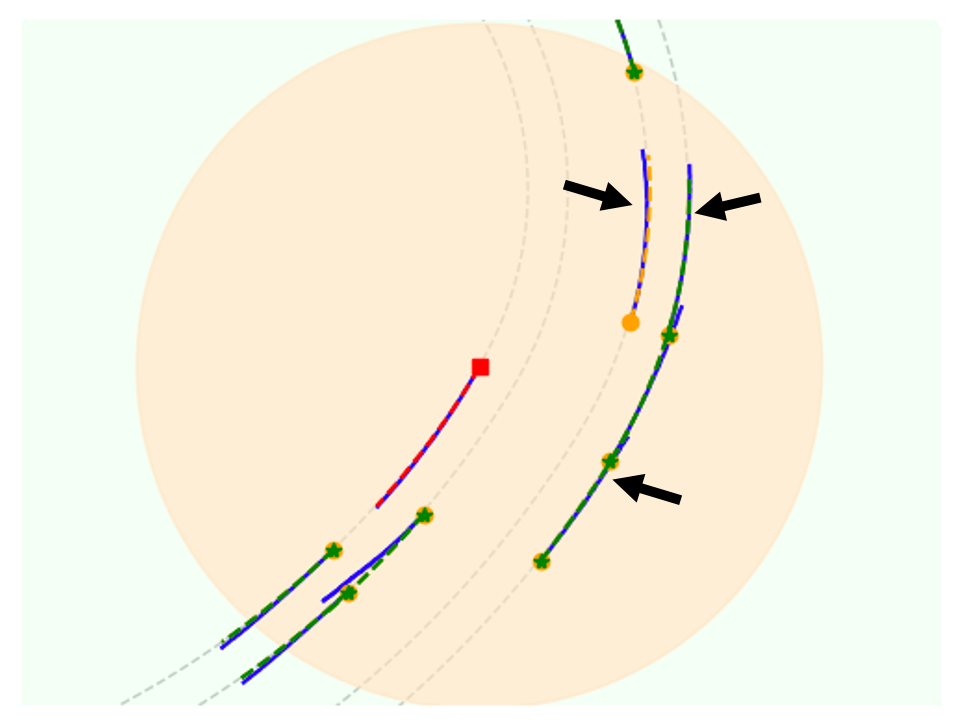}
    \caption{MPR=0.8, ADE=0.36}
  \end{subfigure}
  
  \caption[Visualization of multi-agent prediction at opposite directional turning lanes]{Visualization of multi-agent prediction at opposite directional turning lanes. Same description as in Figure~\ref{figchap4:s480_mpr}.}\label{figchap4:s700_mpr}
\end{figure}

To highlight the extent of improvement, we selected two scenarios for a comparative analysis of multi-agent predictions, specifically under MPR=0 and MPR=0.8. The first scenario includes vehicles navigating a roundabout, as depicted in Figure~\ref{figchap4:s480_mpr}. Notably, when MPR=0, the predictions exhibit a slight deviation, as indicated by the arrows. However, a marked improvement is observed in both vehicles' predictions when they benefit from dual observations under MPR=0.8. Similar improvements are observed in the second scenario shown in Figure~\ref{figchap4:s700_mpr}, where vehicles are driving in opposite directional turning lanes.

\subsection{Comparison with Existing Work}
To evaluate the effectiveness of the proposed architecture MSMA, we compare with the following baselines on the same CARLA dataset:
\begin{itemize}
  \item LSTM Encoder Decoder: We use the vanilla setting with the decoder modified to accommodate 5 modes as described in this work. It predicts the future trajectory for a single agent at a time and only historical trajectory of the agent itself is utilized for prediction. 
  \item VectorNet: We consider the first work using vectorized scene representation described in \cite{gao2020vectornet}. The default settings are employed without auxiliary graph completion. Only the central CAV’s future trajectory gets predicted thus reported. 
  \item HiVT: We use the default setting with 64 hidden dimensions \cite{zhou2022hivt}. All the surrounding agents’ future trajectories were predicted. 
\end{itemize}

Ground truth historical trajectories are used for the baseline models as they do not handle multi-source data. MSMA with 1 frame delay and 0.1 gaussian variance under different MPRs are evaluated. ADEs across all scored agents are calculated. Table~\ref{tabchap4:adecompr} shows the displacement errors at each second and the ADEs for the models being compared. 

We note that LSTM produce the highest ADE as it lacks the information from neighboring agents and map data. While VectorNet shows slightly better performance, it struggles to capture both temporal and spatial relations effectively using a single GNN. On the other hand, HiVT significantly improves trajectory prediction accuracy by explicitly modeling both temporal information and spatial interactions using attention mechanisms. In comparison, MSMA performs even better, particularly when there is more complementary information available from higher CV MPRs.

\begin{table}[t]
    \centering
    \scalebox{0.9}{
    \begin{tabular}{|c|c|c|c|c|c|c|c|}
    \hline
    \multicolumn{2}{|c|}{\multirow{2}{*}{Methods}} &
    \multicolumn{6}{c|}{Performance} \\
    \cline{3-8}
    \multicolumn{2}{|c|}{}&DE@1s & DE@2s & DE@3s &DE@4s& DE@5s &ADE\\
    \hline
    \multirow{5}{*}{MSMA}&MPR=0&0.25 & 0.43 & 0.65 & 0.97 & 1.48 &0.62\\
    \cline{2-8}
    &MPR=0.2&0.24 & 0.43 & 0.64 & 0.95 & 1.47&0.61 \\
    \cline{2-8}
    &MPR=0.4&0.24 & 0.42 & 0.62 & 0.92 & 1.40&0.59 \\
    \cline{2-8}
    &MPR=0.6&0.25 & 0.44 & 0.63 & 0.91 & 1.37&0.59 \\
    \cline{2-8}
    &MPR=0.8&\textbf{0.23} & \textbf{0.40} & \textbf{0.59} & \textbf{0.86} & \textbf{1.33}& \textbf{0.56} \\
    \hline
    \multicolumn{2}{|c|}{LSTM}&0.89 & 0.94 & 1.08 & 1.68 & 3.95 &1.66\\
    \hline
    \multicolumn{2}{|c|}{VectorNet}&0.59 & 1.24 & 1.91 & 2.60 & 3.37 &1.63\\
    \hline
    \multicolumn{2}{|c|}{HiVT}&0.27 & 0.43 & 0.61 & 1.03 & 1.65 &0.66\\
    \hline
    \end{tabular}}
    \caption{Prediction performance comparison with existing works (ADE in meter)}
    \label{tabchap4:adecompr}
\end{table}

\section{Conclusion}
In this work, we proposed a multi-source multi-agent (MSMA) encoder-decoder model for vehicle trajectory prediction in a connected and autonomous vehicle environment. Compared to existing models that rely solely on a single data source, our model incorporates the complementary information provided by both sensors and communication technologies to improve prediction accuracy. A customized CARLA data set was collected, data errors including noise and latency were introduced to accurately reflect the real-world scenarios. A cross-attention fusion module was proposed to align and integrate the perturbed data sources. Numerical experiments showcase that integrating complementary data sources helps improve model prediction performance, particularly when the market penetration rate of connected vehicles is high. 

While our method demonstrates efficacy, it has certain limitations. First, it assumes that CVs broadcast only their own trajectories, overlooking the potential benefits of cooperative driving. Enhanced collaboration could be achieved if each CV possesses information about surrounding vehicles within its communication range, and broadcasting this comprehensive information could augment the surrounding coverage. Second, the model assumes homogeneity among vehicles, and there is room for improvement by considering heterogeneous interactions through graph edges that reflect distinct vehicle characteristics. Last, despite the high-fidelity real-world scenarios provided by CARLA simulation, a real-world dataset would offer deeper insights into vehicle behaviors.

\bibliographystyle{unsrt}  
\bibliography{references}

\begin{thebibliography}{10}

\bibitem{fagnant2015preparing}
Daniel~J Fagnant and Kara Kockelman.
\newblock Preparing a nation for autonomous vehicles: opportunities, barriers and policy recommendations.
\newblock {\em Transportation Research Part A: Policy and Practice}, 77:167--181, 2015.

\bibitem{shladover2018connected}
Steven~E Shladover.
\newblock Connected and automated vehicle systems: Introduction and overview.
\newblock {\em Journal of Intelligent Transportation Systems}, 22(3):190--200, 2018.

\bibitem{vargas2021overview}
Jorge Vargas, Suleiman Alsweiss, Onur Toker, Rahul Razdan, and Joshua Santos.
\newblock An overview of autonomous vehicles sensors and their vulnerability to weather conditions.
\newblock {\em Sensors}, 21(16):5397, 2021.

\bibitem{dosovitskiy2017carla}
Alexey Dosovitskiy, German Ros, Felipe Codevilla, Antonio Lopez, and Vladlen Koltun.
\newblock Carla: An open urban driving simulator.
\newblock In {\em Conference on robot learning}, pages 1--16. PMLR, 2017.

\bibitem{ettinger2021large}
Scott Ettinger, Shuyang Cheng, Benjamin Caine, Chenxi Liu, Hang Zhao, Sabeek Pradhan, Yuning Chai, Ben Sapp, Charles~R Qi, Yin Zhou, et~al.
\newblock Large scale interactive motion forecasting for autonomous driving: The waymo open motion dataset.
\newblock In {\em Proceedings of the IEEE/CVF International Conference on Computer Vision}, pages 9710--9719, 2021.

\bibitem{chang2019argoverse}
Ming-Fang Chang, John Lambert, Patsorn Sangkloy, Jagjeet Singh, Slawomir Bak, Andrew Hartnett, De~Wang, Peter Carr, Simon Lucey, Deva Ramanan, et~al.
\newblock Argoverse: 3d tracking and forecasting with rich maps.
\newblock In {\em Proceedings of the IEEE/CVF Conference on Computer Vision and Pattern Recognition}, pages 8748--8757, 2019.

\bibitem{wilson2021argoverse}
Benjamin Wilson, William Qi, Tanmay Agarwal, John Lambert, Jagjeet Singh, Siddhesh Khandelwal, Bowen Pan, Ratnesh Kumar, Andrew Hartnett, Jhony~Kaesemodel Pontes, et~al.
\newblock Argoverse 2: Next generation datasets for self-driving perception and forecasting.
\newblock In {\em Thirty-fifth Conference on Neural Information Processing Systems Datasets and Benchmarks Track (Round 2)}, 2021.

\bibitem{caesar2020nuscenes}
Holger Caesar, Varun Bankiti, Alex~H Lang, Sourabh Vora, Venice~Erin Liong, Qiang Xu, Anush Krishnan, Yu~Pan, Giancarlo Baldan, and Oscar Beijbom.
\newblock nuscenes: A multimodal dataset for autonomous driving.
\newblock In {\em Proceedings of the IEEE/CVF conference on computer vision and pattern recognition}, pages 11621--11631, 2020.

\bibitem{yu2023v2x}
Haibao Yu, Wenxian Yang, Hongzhi Ruan, Zhenwei Yang, Yingjuan Tang, Xu~Gao, Xin Hao, Yifeng Shi, Yifeng Pan, Ning Sun, et~al.
\newblock V2x-seq: A large-scale sequential dataset for vehicle-infrastructure cooperative perception and forecasting.
\newblock In {\em Proceedings of the IEEE/CVF Conference on Computer Vision and Pattern Recognition}, pages 5486--5495, 2023.

\bibitem{altche2017lstm}
Florent Altch{\'e} and Arnaud de~La~Fortelle.
\newblock An lstm network for highway trajectory prediction.
\newblock In {\em 2017 IEEE 20th international conference on intelligent transportation systems (ITSC)}, pages 353--359. IEEE, 2017.

\bibitem{kim2017probabilistic}
ByeoungDo Kim, Chang~Mook Kang, Jaekyum Kim, Seung~Hi Lee, Chung~Choo Chung, and Jun~Won Choi.
\newblock Probabilistic vehicle trajectory prediction over occupancy grid map via recurrent neural network.
\newblock In {\em 2017 IEEE 20th International Conference on Intelligent Transportation Systems (ITSC)}, pages 399--404. IEEE, 2017.

\bibitem{park2018sequence}
Seong~Hyeon Park, ByeongDo Kim, Chang~Mook Kang, Chung~Choo Chung, and Jun~Won Choi.
\newblock Sequence-to-sequence prediction of vehicle trajectory via lstm encoder-decoder architecture.
\newblock In {\em 2018 IEEE Intelligent Vehicles Symposium (IV)}, pages 1672--1678. IEEE, 2018.

\bibitem{deo2018would}
Nachiket Deo, Akshay Rangesh, and Mohan~M Trivedi.
\newblock How would surround vehicles move? a unified framework for maneuver classification and motion prediction.
\newblock {\em IEEE Transactions on Intelligent Vehicles}, 3(2):129--140, 2018.

\bibitem{lin2021vehicle}
Lei Lin, Weizi Li, Huikun Bi, and Lingqiao Qin.
\newblock Vehicle trajectory prediction using lstms with spatial--temporal attention mechanisms.
\newblock {\em IEEE Intelligent Transportation Systems Magazine}, 14(2):197--208, 2021.

\bibitem{li2019grip++}
Xin Li, Xiaowen Ying, and Mooi~Choo Chuah.
\newblock Grip++: Enhanced graph-based interaction-aware trajectory prediction for autonomous driving.
\newblock {\em arXiv preprint arXiv:1907.07792}, 2019.

\bibitem{cui2019multimodal}
Henggang Cui, Vladan Radosavljevic, Fang-Chieh Chou, Tsung-Han Lin, Thi Nguyen, Tzu-Kuo Huang, Jeff Schneider, and Nemanja Djuric.
\newblock Multimodal trajectory predictions for autonomous driving using deep convolutional networks.
\newblock In {\em 2019 International Conference on Robotics and Automation (ICRA)}, pages 2090--2096. IEEE, 2019.

\bibitem{chai2019multipath}
Yuning Chai, Benjamin Sapp, Mayank Bansal, and Dragomir Anguelov.
\newblock Multipath: Multiple probabilistic anchor trajectory hypotheses for behavior prediction.
\newblock {\em arXiv preprint arXiv:1910.05449}, 2019.

\bibitem{djuric2020uncertainty}
Nemanja Djuric, Vladan Radosavljevic, Henggang Cui, Thi Nguyen, Fang-Chieh Chou, Tsung-Han Lin, Nitin Singh, and Jeff Schneider.
\newblock Uncertainty-aware short-term motion prediction of traffic actors for autonomous driving.
\newblock In {\em Proceedings of the IEEE/CVF Winter Conference on Applications of Computer Vision}, pages 2095--2104, 2020.

\bibitem{zhao2020tnt}
Hang Zhao, Jiyang Gao, Tian Lan, Chen Sun, Benjamin Sapp, Balakrishnan Varadarajan, Yue Shen, Yi~Shen, Yuning Chai, Cordelia Schmid, et~al.
\newblock Tnt: Target-driven trajectory prediction.
\newblock {\em arXiv preprint arXiv:2008.08294}, 2020.

\bibitem{gao2020vectornet}
Jiyang Gao, Chen Sun, Hang Zhao, Yi~Shen, Dragomir Anguelov, Congcong Li, and Cordelia Schmid.
\newblock Vectornet: Encoding hd maps and agent dynamics from vectorized representation.
\newblock In {\em Proceedings of the IEEE/CVF Conference on Computer Vision and Pattern Recognition}, pages 11525--11533, 2020.

\bibitem{khandelwal2020if}
Siddhesh Khandelwal, William Qi, Jagjeet Singh, Andrew Hartnett, and Deva Ramanan.
\newblock What-if motion prediction for autonomous driving.
\newblock {\em arXiv preprint arXiv:2008.10587}, 2020.

\bibitem{varadarajan2022multipath++}
Balakrishnan Varadarajan, Ahmed Hefny, Avikalp Srivastava, Khaled~S Refaat, Nigamaa Nayakanti, Andre Cornman, Kan Chen, Bertrand Douillard, Chi~Pang Lam, Dragomir Anguelov, et~al.
\newblock Multipath++: Efficient information fusion and trajectory aggregation for behavior prediction.
\newblock In {\em 2022 International Conference on Robotics and Automation (ICRA)}, pages 7814--7821. IEEE, 2022.

\bibitem{liang2020learning}
Ming Liang, Bin Yang, Rui Hu, Yun Chen, Renjie Liao, Song Feng, and Raquel Urtasun.
\newblock Learning lane graph representations for motion forecasting.
\newblock In {\em European Conference on Computer Vision}, pages 541--556. Springer, 2020.

\bibitem{zhou2022hivt}
Zikang Zhou, Luyao Ye, Jianping Wang, Kui Wu, and Kejie Lu.
\newblock Hivt: Hierarchical vector transformer for multi-agent motion prediction.
\newblock In {\em Proceedings of the IEEE/CVF Conference on Computer Vision and Pattern Recognition}, pages 8823--8833, 2022.

\bibitem{ye2021tpcn}
Maosheng Ye, Tongyi Cao, and Qifeng Chen.
\newblock Tpcn: Temporal point cloud networks for motion forecasting.
\newblock In {\em Proceedings of the IEEE/CVF Conference on Computer Vision and Pattern Recognition}, pages 11318--11327, 2021.

\bibitem{goli2018vehicle}
Sepideh~Afkhami Goli, Behrouz~H Far, and Abraham~O Fapojuwo.
\newblock Vehicle trajectory prediction with gaussian process regression in connected vehicle environment $\star$.
\newblock In {\em 2018 IEEE Intelligent Vehicles Symposium (IV)}, pages 550--555. IEEE, 2018.

\bibitem{xing2019personalized}
Yang Xing, Chen Lv, and Dongpu Cao.
\newblock Personalized vehicle trajectory prediction based on joint time-series modeling for connected vehicles.
\newblock {\em IEEE Transactions on Vehicular Technology}, 69(2):1341--1352, 2019.

\bibitem{mo2020interaction}
Xiaoyu Mo, Yang Xing, and Chen Lv.
\newblock Interaction-aware trajectory prediction of connected vehicles using cnn-lstm networks.
\newblock In {\em IECON 2020 The 46th Annual Conference of the IEEE Industrial Electronics Society}, pages 5057--5062. IEEE, 2020.

\bibitem{lu2022vehicle}
Yuhuan Lu, Wei Wang, Xiping Hu, Pengpeng Xu, Shengwei Zhou, and Ming Cai.
\newblock Vehicle trajectory prediction in connected environments via heterogeneous context-aware graph convolutional networks.
\newblock {\em IEEE Transactions on Intelligent Transportation Systems}, 2022.

\bibitem{lin2021long}
Lei Lin, Siyuan Gong, Srinivas Peeta, and Xia Wu.
\newblock Long short-term memory-based human-driven vehicle longitudinal trajectory prediction in a connected and autonomous vehicle environment.
\newblock {\em Transportation Research Record}, page 0361198121993471, 2021.

\bibitem{lv2022trajectory}
Pin Lv, Hongbiao Liu, Jia Xu, and Taoshen Li.
\newblock Trajectory prediction with correction mechanism for connected and autonomous vehicles.
\newblock {\em Electronics}, 11(14):2149, 2022.

\bibitem{xu2022v2x}
Runsheng Xu, Hao Xiang, Zhengzhong Tu, Xin Xia, Ming-Hsuan Yang, and Jiaqi Ma.
\newblock V2x-vit: Vehicle-to-everything cooperative perception with vision transformer.
\newblock In {\em European conference on computer vision}, pages 107--124. Springer, 2022.

\bibitem{cui2022coopernaut}
Jiaxun Cui, Hang Qiu, Dian Chen, Peter Stone, and Yuke Zhu.
\newblock Coopernaut: End-to-end driving with cooperative perception for networked vehicles.
\newblock In {\em Proceedings of the IEEE/CVF Conference on Computer Vision and Pattern Recognition}, pages 17252--17262, 2022.

\bibitem{wang2020v2vnet}
Tsun-Hsuan Wang, Sivabalan Manivasagam, Ming Liang, Bin Yang, Wenyuan Zeng, and Raquel Urtasun.
\newblock V2vnet: Vehicle-to-vehicle communication for joint perception and prediction.
\newblock In {\em Computer Vision--ECCV 2020: 16th European Conference, Glasgow, UK, August 23--28, 2020, Proceedings, Part II 16}, pages 605--621. Springer, 2020.

\bibitem{devlin2018bert}
Jacob Devlin, Ming-Wei Chang, Kenton Lee, and Kristina Toutanova.
\newblock Bert: Pre-training of deep bidirectional transformers for language understanding.
\newblock {\em arXiv preprint arXiv:1810.04805}, 2018.

\end{thebibliography}

\end{document}